\documentclass[%
 aip,
 amsmath,amssymb,
 preprint,%
]{revtex4-1}

\usepackage{graphicx}

\usepackage[english]{babel}

\usepackage{natbib}

\usepackage[utf8x]{inputenc}
\usepackage{amsmath}
\usepackage{graphicx}
\usepackage[colorinlistoftodos]{todonotes}

\usepackage{graphicx}
\usepackage{dcolumn}
\usepackage{bm}
\usepackage{xcolor}
\usepackage[normalem]{ulem}
\usepackage[title]{appendix}
\usepackage{comment}
\usepackage{float}
\usepackage{mathtools}

\usepackage{hyperref}

\usepackage{amssymb}
\usepackage{amsthm}
\usepackage{stmaryrd}
\usepackage{amsfonts}
\usepackage{amstext}

\usepackage{subfigure}

\usepackage{epstopdf}
\newcolumntype{C}[1]{>{\centering\arraybackslash$}m{#1}<{$}}
\newlength{\mycolwd}                                
\usepackage{enumitem}

\begin{document}
\title{Reservoir Computing with Noise}
\author{Chad Nathe}
	\affiliation{Mechanical Engineering Department, University of New Mexico, Albuquerque, NM, 87131}
\author{Chandra Pappu}
	\affiliation{Electrical, Computer and Biomedical Engineering Department, Union College, Schenectady, NY, 12309}
\author{Nicholas A. Mecholsky}
	\affiliation{Department of Physics and Vitreous State Laboratory, 
The Catholic University of America,
Washington, DC 20064}
\author{Joseph D. Hart}\affiliation{US Naval Research Laboratory, Washington, DC 20375}
		\author{Thomas Carroll}
	\affiliation{US Naval Research Laboratory, Washington, DC 20375}
\author{Francesco Sorrentino}
	\affiliation{Mechanical Engineering Department, University of New Mexico, Albuquerque, NM, 87131}

\begin{abstract}
    This paper investigates in detail the effects of noise on the performance of reservoir computing. We focus on an application in which reservoir computers are used to learn the relationship between different state variables of a chaotic system.  We recognize that noise can affect differently the training and testing phases. We find that the best performance of the reservoir is achieved when the strength of the noise that affects the input signal in the training phase equals the strength of the noise that affects the input signal in the testing phase. For all the cases we examined, we found that a good remedy to noise is to low-pass filter the input and the training/testing signals;  this typically preserves the performance of the reservoir, while reducing the undesired effects of noise. 
\end{abstract}

\maketitle

\textbf{In any practical application, the noise will inevitably affect the signals that are processed; this is also true in machine learning. Therefore the robustness of a machine learning technique to the noise-corruption of input and training data is an important question. In this work, we investigate the effect of noisy signals on the performance of a reservoir computer acting as an observer. A reservoir observer is an application of reservoir computing in which the internal state of a system is 
reconstructed from knowledge of one or more measured state variables. We find low-pass filtering of the noisy signals that are used by the reservoir observer to be an effective remedy against additive Gaussian noise, provided that the same type of filtering is applied in the training phase and in the testing phase.}

\section{Introduction}

Noise is an unavoidable component in almost all practical applications. For example, signals obtained from biological systems are typically affected by large amount of noise. Hence, it is important to understand how machine learning is affected by noise and what remedies can be put in place to contain its effects.  In this paper, we focus on reservoir computing \cite{jaeger2001echo,maass2002real} as a particular type of machine learning and in particular on reservoir observers, which use knowledge of part of the state of a system to reconstruct the internal state of the same system \cite{lu2017reservoir}. 

{One thing that makes reservoir computers interesting is that they may be implemented as massively parallel devices in analog hardware. This, combined with the simplicity of training, makes them promising for applications such as drones or handheld sensors that require small size, low weight, and low power consumption. Reservoir computers that are all or part analog include photonic systems \cite{appeltant2011,larger2012, van_der_sande2017, hart2019,chembo2019,argyris2019}, analog electronic circuits \cite{schurmann2004}, mechanical systems \cite{dion2018} and field programmable gate arrays \cite{canaday2018}. Many other examples are included in the review paper \cite{tanaka2019}. }

A number of authors have considered how noise added to the input, testing, or training signals affect the performance of reservoir computing. In Reference \onlinecite{jungling2022} the authors used the concept of consistency to show how added noise decreased the information processing capacity of reservoir computers. Vettelschoss et al. \cite{vettelschoss2022} demonstrated the change in information processing capacity in a single-node reservoir computer. Shougat et al. \cite{shougat2021} added noise to the input mask for a reservoir computer based on a Hopf oscillator. References \onlinecite{carroll2018a} and \onlinecite{carroll2022b} examined the effect of added noise on the ability to classify different chaotic signals.  In Reference \onlinecite{liao2021} the authors sought to mitigate the effects of added noise by using a bistable function for the activation function to take advantage of the principle of stochastic resonance. Reference \onlinecite{lu2017reservoir} 
studied the effects of noise on the input signal for the case of an observer reservoir. Reference \onlinecite{pathak2017using} 
investigated the impact of measurement noise on the input data for Lyapunov exponent estimation using reservoir computing. Reference \cite{semenova2019fundamental} studied the effects of noise on signal-to-noise ratio in reservoir computers with no nonlinearity. In References \onlinecite{estebanez2019, rohm2021, kong2021} noise occupied a constructive role in attractor reconstruction tasks. In these papers, noise added in the training stage allowed the reservoir computer to learn behaviors in parameter ranges that were not part of the training data.
A recent paper \cite{donati2022noise} has investigated the effects of noise in an experimental reservoir computer.

In this work, we study how measurement noise added to the reservoir input and/or output signals affects the reservoir computer performance for the so-called observer task \cite{lu2017reservoir}. This would be the situation when using the reservoir computer to model any real-world system since the input data will be corrupted by some amount of measurement noise. 
We also consider the general situation in which the noise strength in the input and output signals can be different in the training and testing phases, as would be the case when training is done in the lab and the reservoir is deployed in the field. 

We find that the reservoir performs best when the measurement noise on the input signal in the training and testing phases has the same strength. Since it may not always be possible to match the noise in the training phase to the noise in the testing phase, we propose applying a simple low-pass filter that is well-matched to the true signal spectrum to the input signal. We find that such a filter significantly mitigates the negative effect of the noise.

The rest of this paper is organized as follows. In Sec.\ II, we introduce the reservoir equations. In Sec.\ III we describe the effects of noise on the training error and the testing error. In Sec.\ IV, we introduce low-pass filtering as a remedy to noise. Finally, the conclusions are given in Sec.\ V.


\section{Reservoir Computing in the presence of noise}

\begin{figure}[!ht]
     \centering
     \includegraphics[width=6.60in]{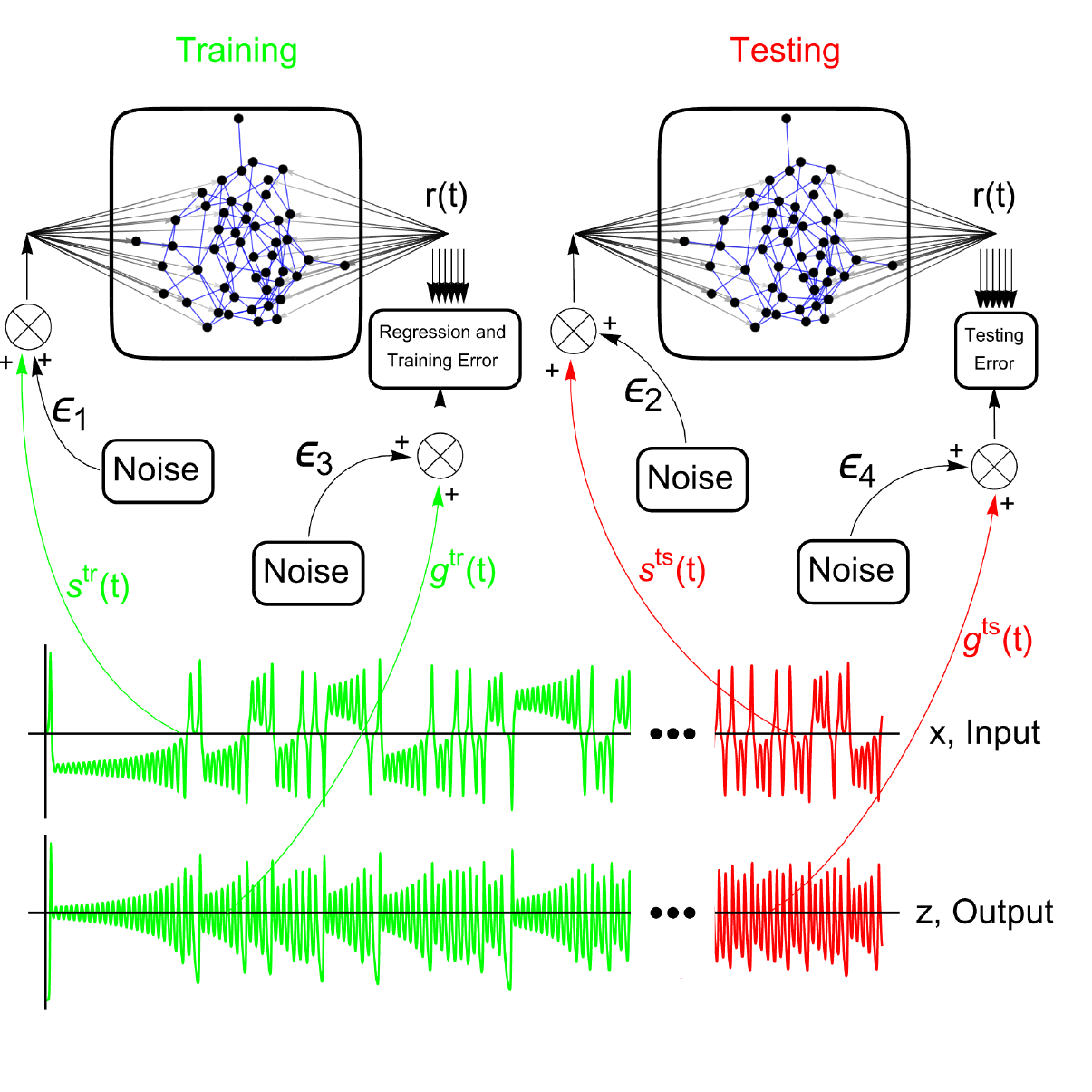}
     \caption{Illustration of how noise can affect the input and output signals of a reservoir computer in the training and testing phase. {We are performing an ``observer task'' in which we feed the $x$ coordinate of the Lorenz system started from random initial conditions into the reservoir and the reservoir is trained to predict the $z$ coordinate}. We call $\epsilon_1$ the strength of the noise affecting the input signal in the training phase, $\epsilon_2$ the strength of the noise affecting the input signal in the testing phase, $\epsilon_3$ the strength of the noise affecting the training signal, and $\epsilon_4$ the strength of the noise affecting the testing signal.} \label{figmain}
\end{figure}

{
In this section, we present a general formulation of the reservoir dynamics in the presence of noise. A signal from the underlying system to be observed is used to drive the reservoir (the input signal), and the reservoir is trained to reproduce a second signal from the underlying system (the training signal). However, in our formulation, both the input and training signals are affected by noise. In particular,}
 we consider the case of additive white Gaussian noise (AWGN) applied to the reservoir input and output. We call $\epsilon_1 \geq 0$ the noise strength added to the input signal in the training phase, $\epsilon_2 \geq 0$ the strength of the noise added to the input signal in the testing phase, $\epsilon_3 \geq 0$ the strength of the noise added to the training signal, and $\epsilon_4 \geq 0$ the strength of the noise added to the testing signal. {All of these noise sources are independent and identically distributed (iid).} This is illustrated in Fig.\ \ref{figmain} which shows how noise can affect either the input or the output signals that interact with the reservoir. The motivation for assuming $\epsilon_1 \neq \epsilon_2$ and $\epsilon_3 \neq \epsilon_4$ is to account for situations in which training is performed in a laboratory and testing in the field. It is thus expected that the level of noise may vary substantially between the training and the testing phases.

\onecolumngrid

In the rest of this paper, for a given noise-free signal $x(t)$ we will use the following notation: $\tilde{x}_\epsilon(t)$
is the noise corrupted version of $x(t)$ with $\epsilon \geq 0$ being the noise strength and $\hat{x}_\epsilon(t)$ is the low-pass filtered (LPF) version of $\tilde{x}_\epsilon(t)$. 
We model the reservoir dynamics in discrete time, while the underlying physical system with which the reservoir interacts evolves in continuous time.
Thus input and output signals are  sampled at each time step of the reservoir dynamics, with sampling period $t_s$. {We normalize the noise-free input and output signals so that their mean is equal to zero and their standard deviation is equal to one.} We then set the normalized noise-corrupted signal $\tilde{x}_\epsilon(t)=x(t) +\epsilon \sqrt{\frac{t_s}{T}} \zeta(t)$, where at each discrete time $t$, $\zeta(t)$ is a scalar drawn from a standard normal distribution and $T$ is the approximate period over which the  signal completes one oscillation. {The weighting $\sqrt{t_s/T}$ is to renormalize the standard deviation of the sum of $T/t_s$ standard normal values added to the reservoir \cite{sorrentino2009using}.} 
Note that by taking the noise-free signal $x(t)$ to have mean equal zero and standard deviation equal one, the noise strength $\epsilon$ can be directly translated into a measure of signal-to-noise ratio (SNR), that is, $\mbox{SNR}={T}/{(t_s \epsilon^2)}.$

The equation that models the reservoir dynamics is,

\begin{equation} \label{restr}
{\mathbf{r}}(t+1) = (1-\alpha)\mathbf{r}(t) + \alpha \tanh(A\mathbf{r}(t) + \mathbf{w} \tilde{s}_{\epsilon_1}^{tr}(t))
\end{equation}
{where, $\alpha$ is the leakage rate chosen in the range of $[0,1]$,} $A$ is the coupling matrix, 
$\tilde{s}_{\epsilon_1}^{tr}(t)$ is the noise-corrupted input signal in the training phase, and $\mathbf{w}$ is a vector of random elements drawn from a Gaussian distribution with mean $1$ and standard deviation $0.1$, i.e., $\mathcal{N}(0.01,1)$. The matrix $A$ is the adjacency matrix of an undirected and unweighted Erdos Renyi network  with $N=100$ nodes and connectivity probability $p=0.5$. We set the elements on the main diagonal to be equal to zero; then normalize the matrix $A \leftarrow A/ \rho(A)$, where $\rho(A)$ is the spectral radius so that the largest  eigenvalue of the normalized matrix has modulus equal to $1$. {Overall, we found our results that follow to not be strongly affected by the particular choice of the network topology, see Sec.\ III of the Supplementary Information for a study of the effects of the network topology.}

{ In this paper we consider three different tasks (described in detail in the Supplementary Information Sec.\ SI), which we briefly refer to as the Lorenz task, the Rossler task, and the Hindmarsh Rose (HR) task. We optimize the parameter $\alpha$ with respect to the particular task assigned to the reservoir and set $\alpha=0.1$ for the Lorenz task, $\alpha = 0.003$ for the Rossler task, and $\alpha=0.01$ for the HR task. Further information on the optimization in $\alpha$ is presented in the Supplementary Information Sec.\ SIV.}

From the solution of Eq.\ \eqref{restr} one obtains the readout matrix,
\begin{equation}
\Omega=
\left(\begin{array}{ccccc}
    r_1(1) & r_2(1) & ... & r_N(1) & 1\\
    r_1(2) & r_2(2) & ... & r_N(2) & 1\\
    \vdots & \vdots & \vdots &  \vdots  & \\
    r_1(T_1) & r_2(T_1) & ... & r_N(T_1) & 1\\
\end{array}\right)
\end{equation}
where $r_i(t)$ is the readout of node $i$ at time $t$ and $t=T_1$ indicates the end of the training phase. The last column of $\Omega$ is set to 1 to account for any constant offset in the fit. We then relate the readouts to the training signal, $\mathbf{g}^{tr}$, with additive noise, via the unknown coefficients contained in the vector, $\pmb{\kappa}$, 

\begin{equation}
    \Omega \pmb{\kappa} = {\tilde{\mathbf{g}}_{\epsilon_3}^{tr}} 
\end{equation}
where ${\tilde{\mathbf{g}}_{\epsilon_3}^{tr}}$ is the noisy training signal. We then compute the unknown coefficients vector $\pmb{\kappa}$ via the equation,

\begin{equation} \label{kappa}
   \pmb{\kappa} =  \mathbf{\Omega}^{\dagger}{\tilde{\mathbf{g}}_{\epsilon_3}^{tr}}.
\end{equation}

{Here, $\mathbf{\Omega}^{\dagger}$ is given as}

\begin{equation} \label{kappa}
   \mathbf{\Omega}^{\dagger} =  \mathbf{\left(\Omega^T \Omega + \beta I \right)^{-1}\Omega^T}.
\end{equation}
{In the above equation, $\beta$ is the ridge-regression parameter used to avoid overfitting \cite{lu2017reservoir} and $\mathbf{I}$ is the identity matrix.} Next, we define the training fit signal as
\begin{equation}
\mathbf{h} = \Omega \pmb{\kappa}.
\end{equation}

Lastly, the training error is computed as,
\begin{equation} \label{tre}
\Delta_{\text{tr}} = \frac{\langle\mathbf{h} - {\tilde{\mathbf{g}}_{\epsilon_3}^{tr}} \rangle}{\langle{\tilde{\mathbf{g}}_{\epsilon_3}^{tr}} \rangle}
\end{equation}
where the notation $\langle \rangle$ denotes the standard deviation.

In the testing phase, the reservoir evolves according to the equation,
\begin{equation} \label{reste}
{\mathbf{r}}(t+1) = (1-\alpha)\mathbf{r}(t) + \alpha \tanh(A\mathbf{r}(t) + \mathbf{w} \tilde{s}_{\epsilon_2}^{ts}),
\end{equation}
where $\tilde{s}_{\epsilon_2}^{ts}$ is the noise-corrupted version of the input signal in the testing phase. Typically in our numerical experiments, we have the testing phase follow immediately after the training phase (but this is not a requirement). Similarly to what was done in the training phase, we compute the matrix,
\begin{equation}
\check{\Omega}=
\left(\begin{array}{ccccc}
    r_1(1) & r_2(1) & ... & r_N(1)  & 1\\
    r_1(2) & r_2(2) & ... & r_N(2) & 1\\
    \vdots & \vdots & \vdots &  \vdots  & \\
    r_1(T_2) & r_2(T_2) & ... & r_N(T_2) & 1\\
\end{array}\right)
\end{equation}
where $T_2$ indicates the end of the testing phase and we typically set, $T_2 = 
\frac{1}{2}T_1$. We then compute the testing fit signal by the equation, 
\begin{equation}\label{testing_fit}
\check{\mathbf{h}} = \check{\Omega} \pmb{\kappa} 
\end{equation}
where the vector $\pmb{\kappa}$ is the one obtained  in the training phase (Eq.\ \eqref{kappa}). The testing error is equal to,
\begin{equation} \label{tse}
\Delta_{\text{ts}} = \frac{\langle\check{\mathbf{h}} - \tilde{\mathbf{g}}_{\epsilon_4}^{ts}\rangle}{\langle\tilde{\mathbf{g}}_{\epsilon_4}^{ts}\rangle}
\end{equation}
where $\tilde{\mathbf{g}}_{\epsilon_4}^{ts}$ is the noise-corrupted the testing signal.

\begin{figure} [H]
 \centering
 \includegraphics[scale=0.65]{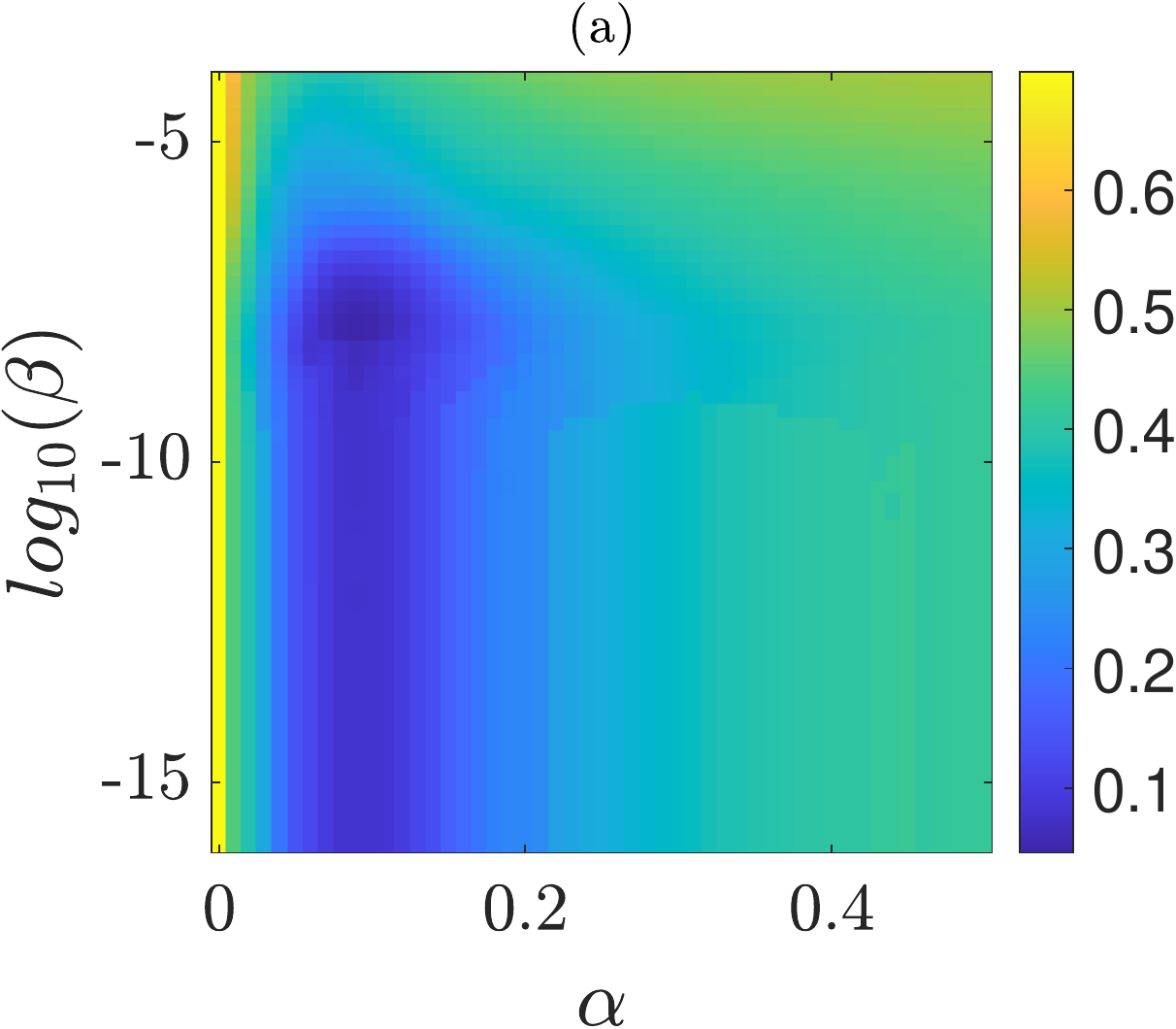}           
  \includegraphics[scale=0.65]{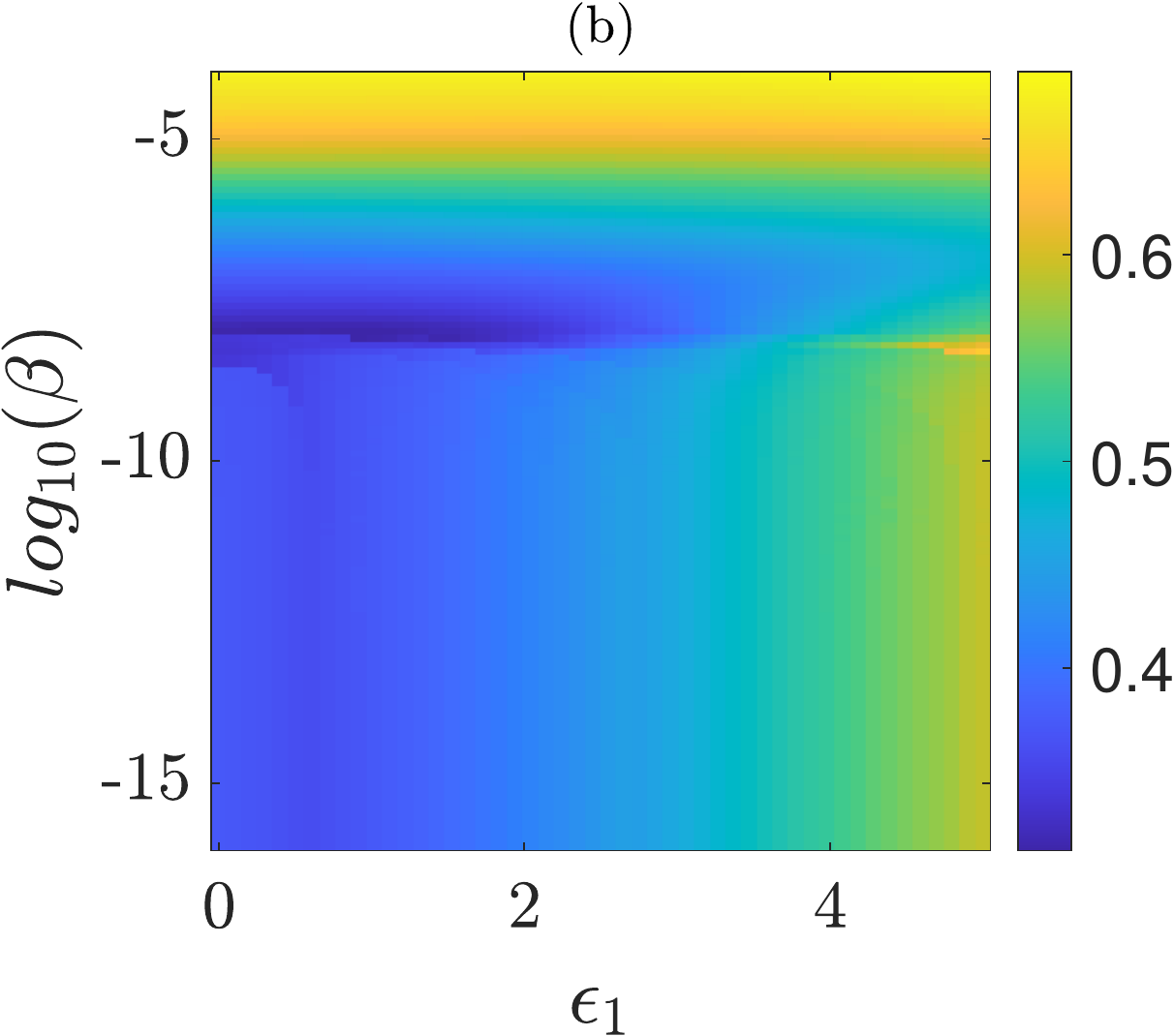}
 \caption{Contour plot of the testing error for the Lorenz system, as a function of (a) $\alpha$ and $\beta$, for a case in which  noise is absent from the training and testing phases and (b)  as a function of $\epsilon_1$ and $\beta$. Here an optimized value of $\alpha=0.1$ is considered. }
 \label{fig:alpha_beta_optimization}
\end{figure}

In the figures that follow, we average several simulations over the choice of the matrix $A$ and over different noise realizations. Figure \ref{fig:alpha_beta_optimization} illustrates our  selection of the hyperparameters $\alpha$ and $\beta$ of the reservoir computer. First, we computed the testing error as a function of  the leakage rate $\alpha$ and of the ridge-regression parameter $\beta$. As can be seen from Fig.\ \ref{fig:alpha_beta_optimization}(a), in the absence of noise, the minimum testing error is obtained when $\alpha = 0.1$ and $\beta = 10^{-8}$, approximately. We then set $\alpha$ equal to the optimum value $0.1$ and compute the testing error as a function
of both $\beta$ and $\epsilon_1$, i.e., the noise strength
added to the input signal in the training phase, which is shown in Fig.\ \ref{fig:alpha_beta_optimization}(b). We see that
as $\epsilon_1$ is varied, the minimum testing error is always obtained when $\beta$ is around $10^{−8}$. Therefore, in what follows, we use the optimal values $\alpha=0.1$ and $\beta=10^{-8}$. 
We conducted further simulations and verified that other choices of $\beta$ such as $10^{−9}$ or $10^{−7}$ still produce similar qualitative results, 
but with only a slight change in the magnitude of the error. We conclude that our proposed methodology is not too sensitive to the particular choice of the hyperparameter $\beta$.

\section{Testing in the presence of noise}

\begin{figure}[H]
    \centering

    \includegraphics[scale=1.25]{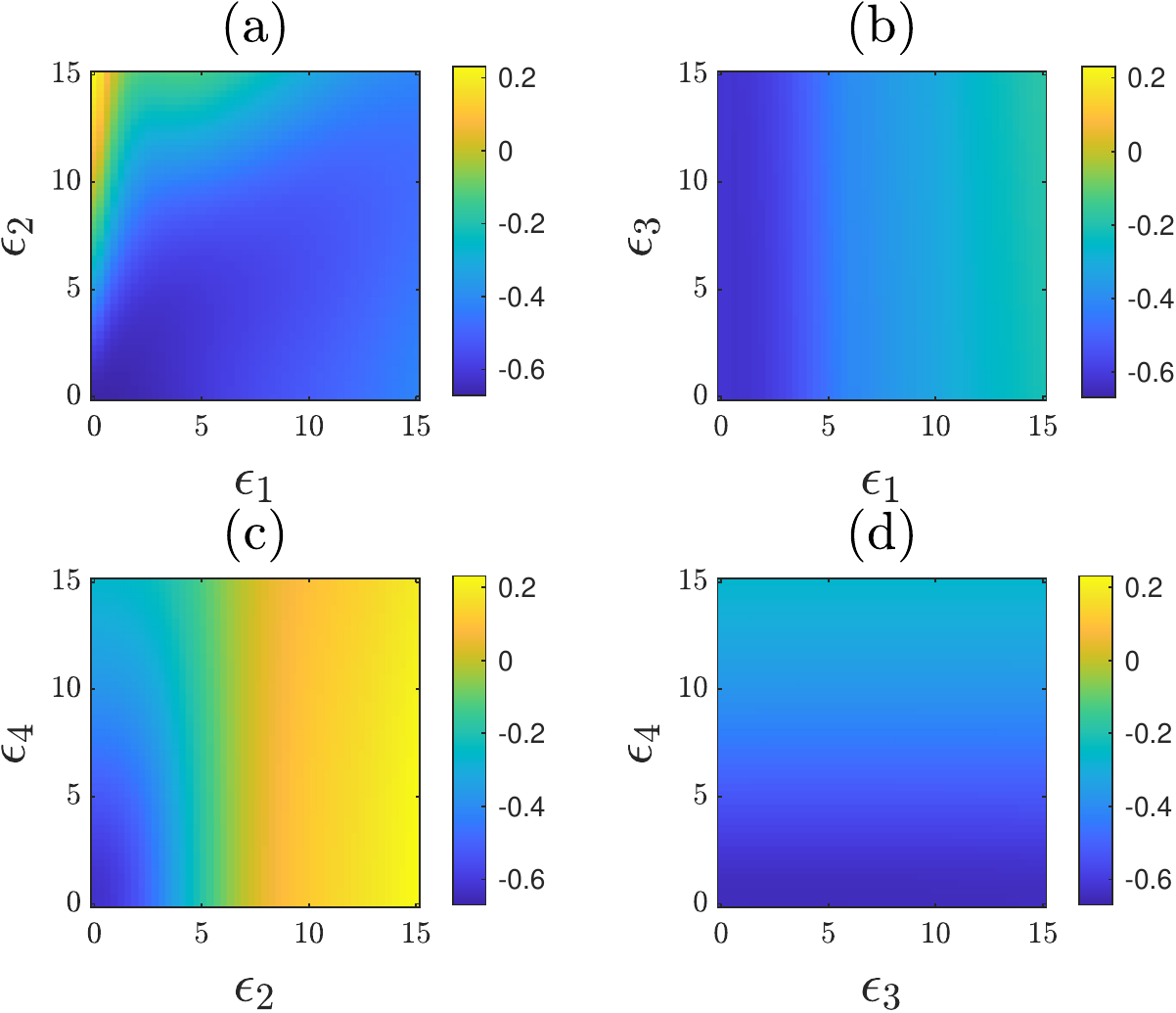}
    
    \caption{Contour plots of the testing error for the Lorenz system, when (a)  $\epsilon_1$ and $\epsilon_2$ are varied while $\epsilon_3=\epsilon_4 = 0$, (b) $\epsilon_1$ and $\epsilon_3$ are varied while $\epsilon_2=\epsilon_4 = 0$, (c) $\epsilon_2$ and $\epsilon_4$ are varied while $\epsilon_1=\epsilon_3 = 0$, and (d) $\epsilon_3$ and $\epsilon_4$ are varied while $\epsilon_1=\epsilon_3 = 0$. For illustrative purposes, we depicted the error as $\log_{10}$ of the actual error.}
     \label{fig:contours}
\end{figure}

We show in Fig.\ \ref{fig:contours}  contour-level plots of the testing error $\Delta_{ts}$ as we vary $\epsilon_1$ and $\epsilon_2$ in (a),  $\epsilon_1$ and $\epsilon_3$ in (b), $\epsilon_2$ and $\epsilon_4$ in (c) and $\epsilon_3$ and $\epsilon_4$ in (d). In Fig.\  \ref{fig:contours}(a), the noise strengths of the training input signal $\epsilon_1$ and testing input signal $\epsilon_2$ are varied while keeping $\epsilon_3=\epsilon_4=0$. Optimal performance is achieved when both $\epsilon_1$ and $\epsilon_2$ are low, as expected. It is quite natural that training and testing should be performed under the same environmental conditions. Therefore, the performance is quite good when $\epsilon_1\approx\epsilon_2$.

When $\epsilon_1$ is small, and $\epsilon_2$ is large (corresponding to a small input noise while training and a large input noise while testing), the testing error $\Delta_{ts}$ is very high (e.g., $\epsilon_1<5$ and $\epsilon_2>10$). On the other hand, when the reservoir is trained with highly noisy signals and the testing input signal has lower noise levels (e.g., $\epsilon_1>10$ and $\epsilon_2<5$), the testing error is reduced by an order of magnitude. The important takeaway from Fig.\  \ref{fig:contours}(a) is that it is best to train a reservoir computer on an input signal with the same amount of noise as will be present during operation. In situations where an estimate of the noise that will be present during operation is difficult to obtain ahead of time, it is much better to train a reservoir computer on an input signal with too much noise (compared to what it will receive in operation) than with too little.

In Fig.\  \ref{fig:contours}(b), $\epsilon_1$ and $\epsilon_3$ (the noise strength on the training signal $g^{tr}(t)$) are varied. In this case $\epsilon_2 = \epsilon_4 = 0$ is considered. The reservoir computer is trained with input $\tilde{s}_{\epsilon_1}^{tr}(t)$, which affects the reservoir dynamics $\mathbf{r}(t)$. For a large $\epsilon_1$, $\tilde{s}_{\epsilon_1}^{tr}(t)$ has an increased amplitude and generates a qualitatively different response in the nonlinear reservoir than for the case of small $\epsilon_1$. Since the testing signal does not have any noise, the $\Delta_{ts}$ proportionally increase to $\epsilon_1$. On the other hand, from Eq. \ref{testing_fit}, fitting a training signal is identical to averaging the uncorrelated noise from each reservoir node. Hence, the $\Delta_{ts}$ is quite robust to change in $\epsilon_3$.

In Fig.\  \ref{fig:contours}(c), $\epsilon_2$ and $\epsilon_4$ (the noise strength on the testing signal $g^{ts}(t)$) are varied and $\epsilon_1 = \epsilon_3$ is considered as 0. From Eq. \ref{reste}, the reservoir dynamics in the testing phase are determined by $\tilde{s}_{\epsilon_2}^{ts}$ and are altered as $\epsilon_2$ changes. However, $\tilde{g}^{ts}_{\epsilon_4}(t)$ appears only in the evaluation of the testing error Eq.~\ref{tse} and does not affect the reservoir dynamics. Therefore, $\epsilon_2$ dominates the testing error for higher values of $\epsilon_2$. An important takeaway from Figs.~ \ref{fig:contours}(b)-(c) is that noise on the reservoir input signal can be much more problematic than noise on the training or testing output signals.

In Fig.~ \ref{fig:contours}(d), the testing error is computed as a function of $\epsilon_3$ and $\epsilon_4$ while assuming $\epsilon_1=\epsilon_2 = 0$. Similar to Fig.~ \ref{fig:contours}(b), the testing error $\Delta_{ts}$ is robust to $\epsilon_3$, due to the averaging in the training stage. However, with increase in $\epsilon_4$, the testing error $\Delta_{ts}$, which is function of $\tilde{g}^{ts}_{\epsilon_4}(t)$, increases linearly.

    

\section{Low-Pass Filtering}

In this section, in order to improve the performance of reservoir computing in the presence of noise, the noise-corrupted versions of the input signal ($s(t)$) and of the output signal ($g(t)$) are driven through a low pass filter (LPF). The LPF rejects high-frequency components while allowing the frequencies below the chosen cutoff frequency. The equation of a first-order LPF is 
\begin{align}
    \frac{dV_{out}}{dt}&=\frac{1}{\tau}\left[V_{in}-V_{out}\right] \mbox{, or equivalently,}\nonumber\\\\
  \dot{\hat{x}} &=a\left[\tilde{x}-\hat{x}\right]\nonumber
\end{align}
where $a=\frac{1}{\tau}$ is the cutoff frequency, $\tilde{x}(t)=V_{in}(t)$ is the LPF input and $\hat{x}(t)=V_{out}(t)$ is the LPF output. We chose this type of filter because it is characterized by only a single parameter, and therefore reduces the complexity of the analysis and any potential physical implementations. 

Adjusting the reservoir parameters such as $\alpha$ will alter the bandpass characteristics of the reservoir computer, but because $\alpha$ multiplies a nonlinear function on the right hand side of Eq.\ \eqref{restr}, adjusting $\alpha$ to alter the reservoir bandwidth will also change the reservoir nonlinearity so that it is no longer optimized for reproducing the Lorenz chaotic signal. Adding a separate linear low pass filter allows us to alter the bandwidth of the reservoir computer without affecting its nonlinear characteristics.

This noise reduction method is particularly attractive because, in a field-deployed reservoir computer, low-pass filters are straightforward to implement, either with analog components or digital signal processing. In some types of photonic reservoir computers such as optoelectronic oscillators, a filter may even be directly integrated as part of the reservoir computer itself \cite{dai2021classification,dai2022rf}. Low-pass filters have previously been used on the individual node outputs to expand the reservoir \cite{carroll2021adding}, but to our knowledge, this is the first comprehensive investigation of the use of a low-pass filter to mitigate the effects of noise on reservoir computing performance.

If the input signals are low-pass filtered, Eq.\ \eqref{restr} (Eq.\ \eqref{reste}) are evolved with $\tilde{s}_{\epsilon_1}^{tr}(t)$ replaced by $\hat{s}_{\epsilon_1}^{tr}(t)$ ($\tilde{s}_{\epsilon_2}^{ts}(t)$ replaced by $\hat{s}_{\epsilon_2}^{ts}(t)$). If the output signals are low-pass filtered,  $\tilde{g}_{\epsilon_3}^{tr}(t)$ is replaced by  $\hat{g}_{\epsilon_3}^{tr}(t)$ in Eqs.\ \eqref{kappa} and \eqref{tre} ( $\tilde{g}_{\epsilon_4}^{ts}(t)$ is replaced by $\hat{g}_{\epsilon_4}^{ts}(t)$ in Eq.\ \eqref{tse}.)

{We first consider an ideal situation in which the underlying system is known a \textit{priori}; then using statistical analysis, the optimal cutoff frequency $\widehat{a}$ for the LPF can be determined}. {The optimal cutoff frequency $\widehat{a}$ can be 
obtained from knowledge of the spectrum of the input signal, which is}
{\begin{align}
    X(f)&= |\mathcal{F}[x(t)]|\nonumber
\end{align}
where, $\mathcal{F}$ is the Fourier transform operator.} \\

\begin{figure} [H]
 \centering
 \includegraphics[scale=0.4]{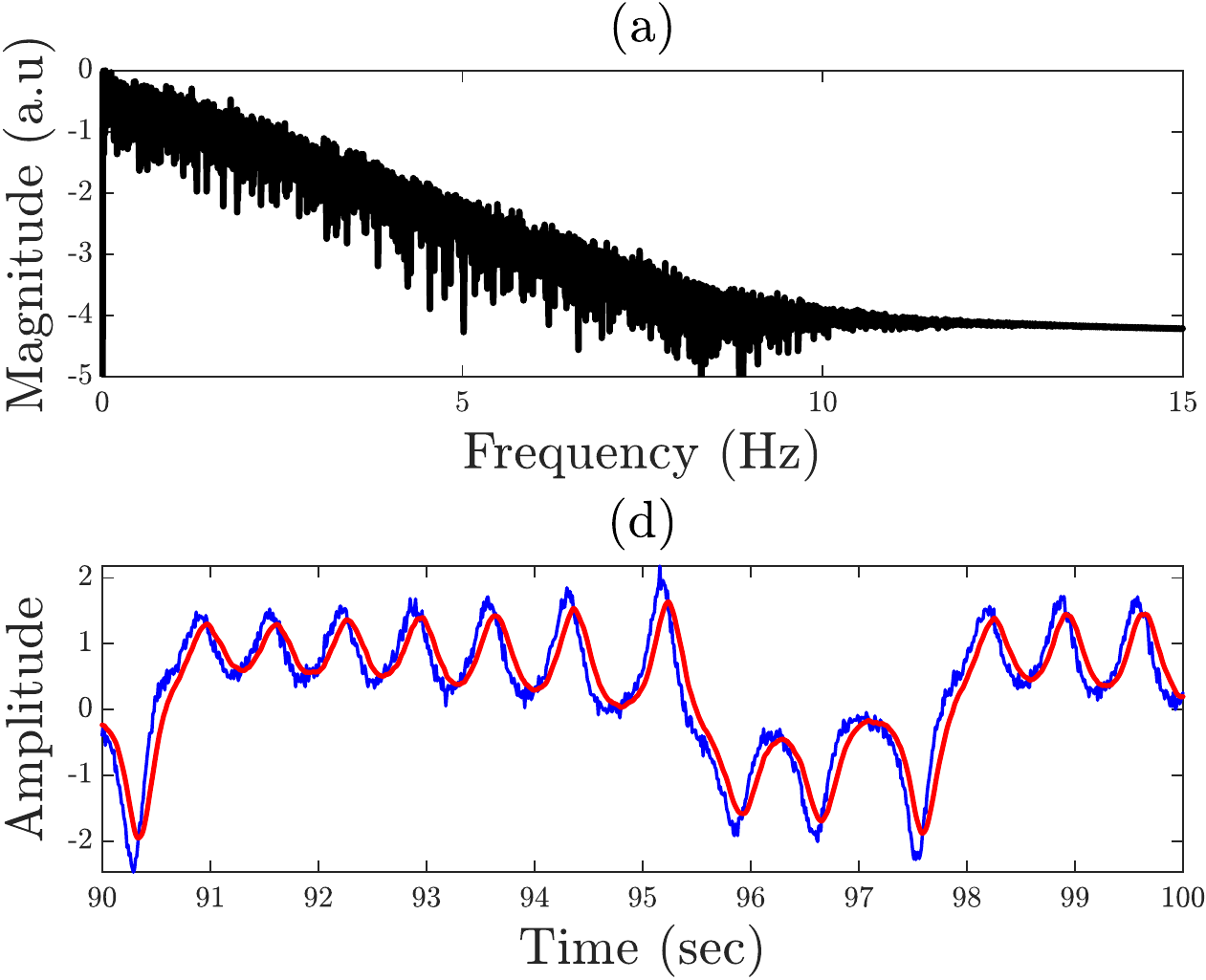}
 \includegraphics[scale=0.4]{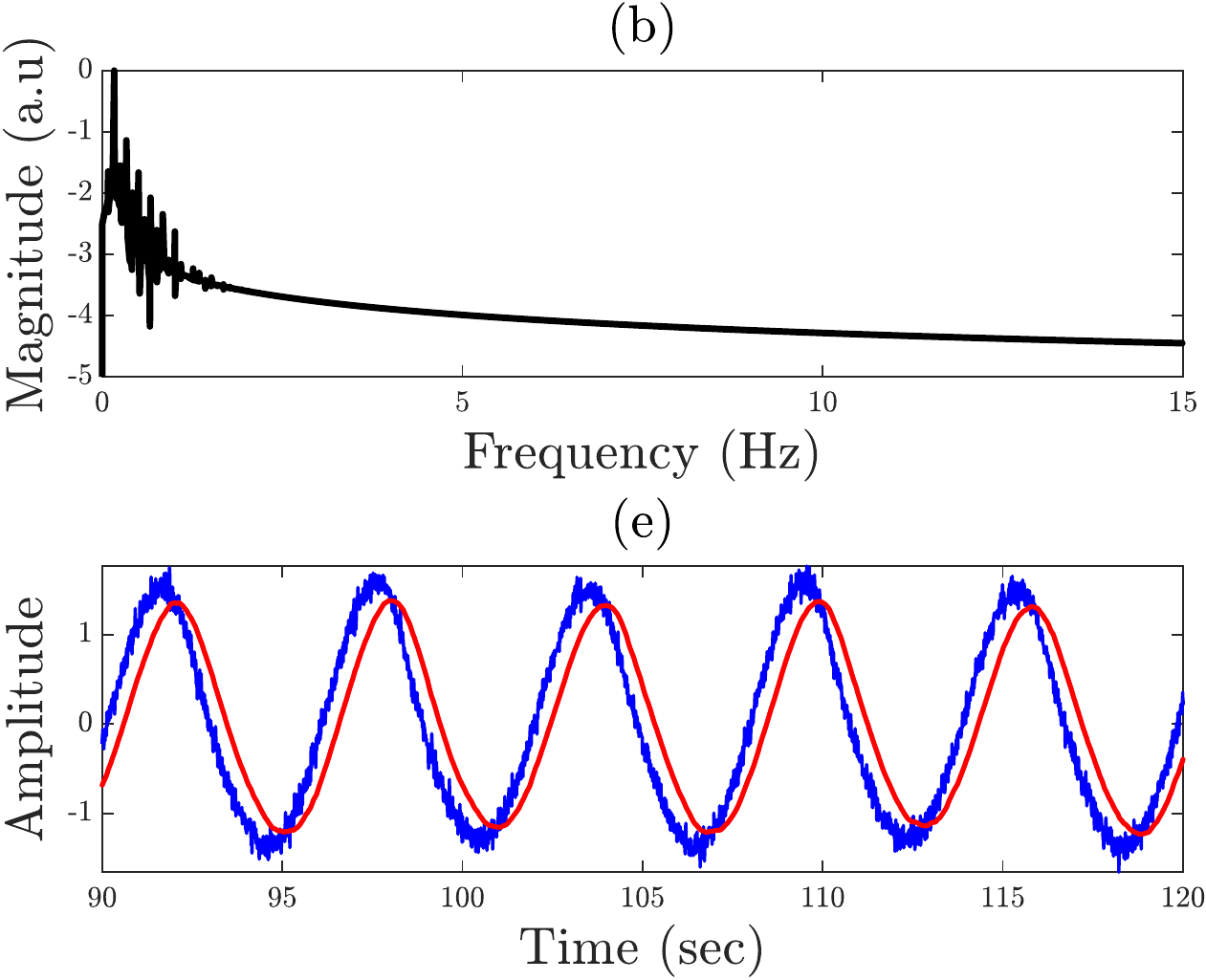}
 \includegraphics[scale=0.4]{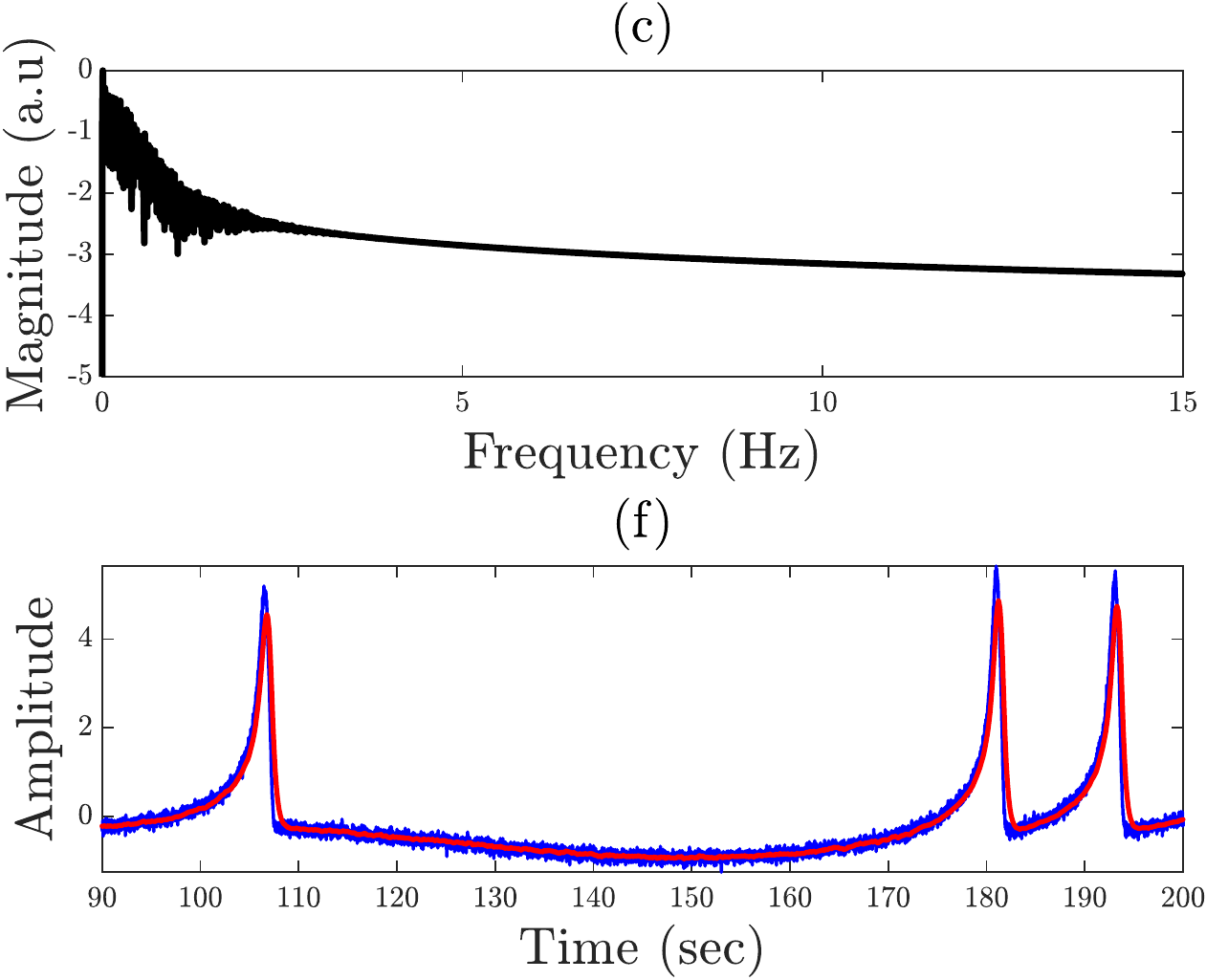}
 \caption{The top panels show the spectrum of state variable $x(t)$ for the (a) Lorenz, (b) Roessler, and (c) Hindmarsh-Rose systems. The bottom panels show the time series plots of $\tilde{x}(t)$ (blue color) and $\hat{x}(t)$ (red color) for the (a) Lorenz, (b) Roessler, and (c) Hindmarsh-Rose systems.}
 \label{fig:Spectrum}
\end{figure}

{Figure \ref{fig:Spectrum} shows the spectrum of the input signal for the cases of the Lorenz, Roessler and Hindmarsh-Rose systems. The signals generated from chaotic systems have an invariant density function. Consequently, the shape of the spectrum of $x(t)$ is not altered even by changing the system's initial conditions. That is, for multiple chaotic signal realizations, their spectrum shape is invariant.} The spectrum plot shown in Figure \ref{fig:Spectrum} is generated from one such realization.

{From Fig.\ \eqref{fig:Spectrum}, we observe that the maximum frequency component of $x(t)$ for the Lorenz system is at around 12 Hz. Similarly, the highest frequency for the Roessler system is at 2 Hz, and that of the Hindmarsh-Rose system is at 3 Hz. Therefore, the values of $\widehat{a}$ for the Lorenz, Roessler, and Hindmarsh-Rose systems are 12, 2, and 3, respectively. If the cutoff frequency exceeds the optimal value, the filter allows more noise to go through. On the other hand, if it is less than the optimal value, it filters out the chaotic signal $x(t)$. Consequently, the minimum error is expected at $\widehat{a}$, and the error increases when the choice of $a$ is not optimal. The lower plots show the noise-corrupted version of the drive signal with $\epsilon_1=2$ and $\epsilon_2=5$ in blue and the filtered version of the input signal in red. It can be observed that $\hat{x}(t)$ appears to be delayed with respect to $\hat{x}(t)$, while the noise is filtered.} 

\begin{figure} [H]
 \centering
 \includegraphics[scale=0.4]{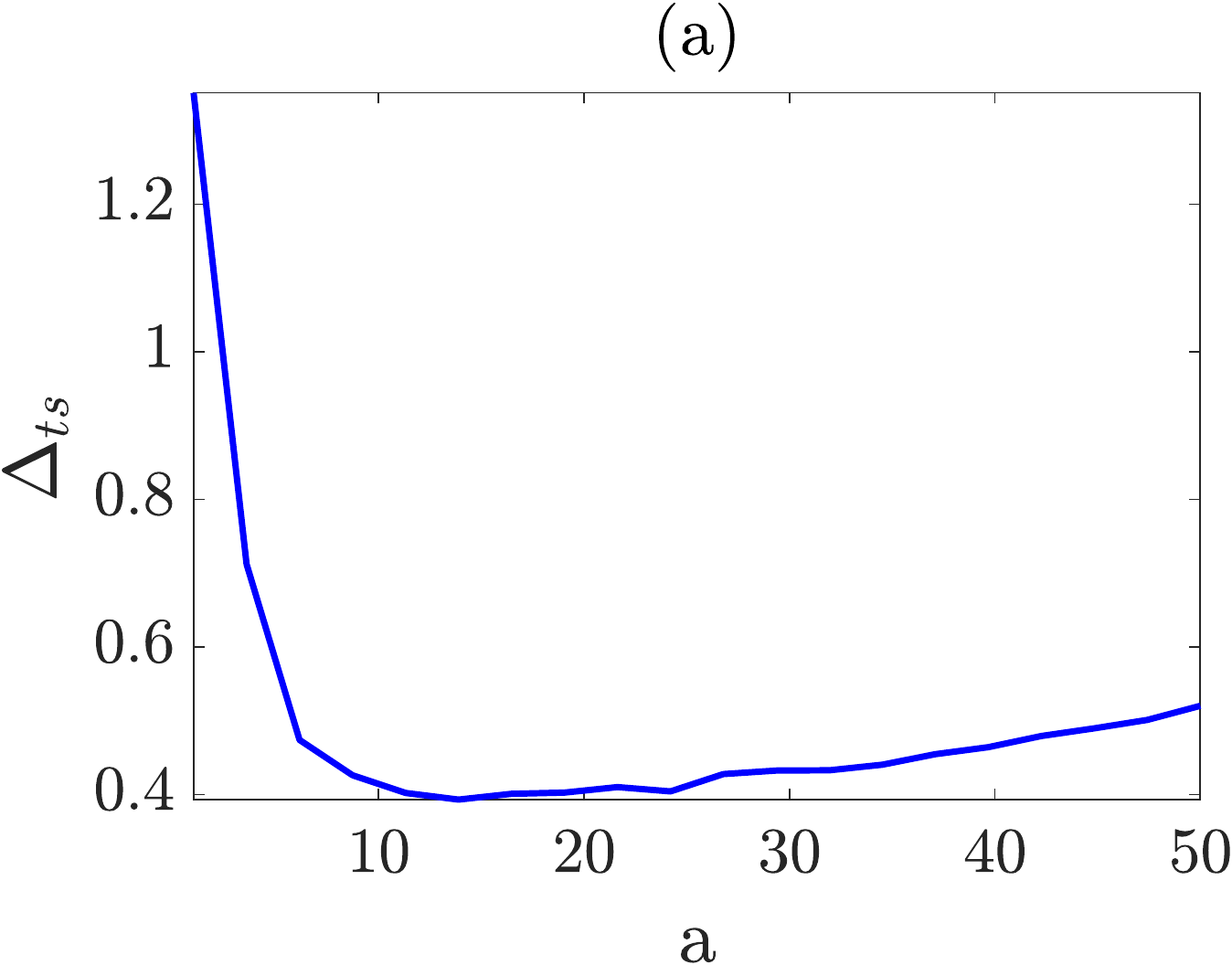}
 \includegraphics[scale=0.4]{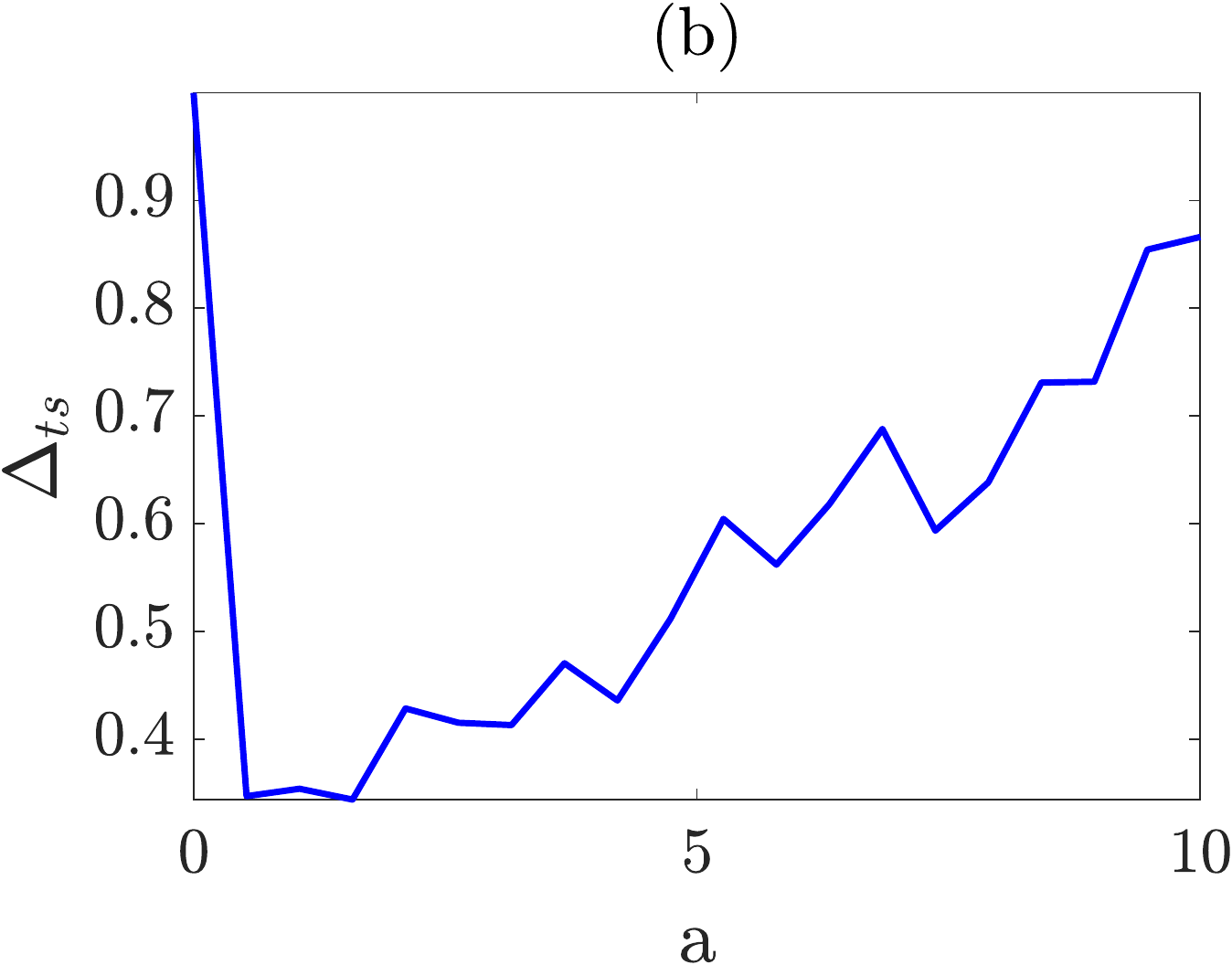}
\includegraphics[scale=0.4]{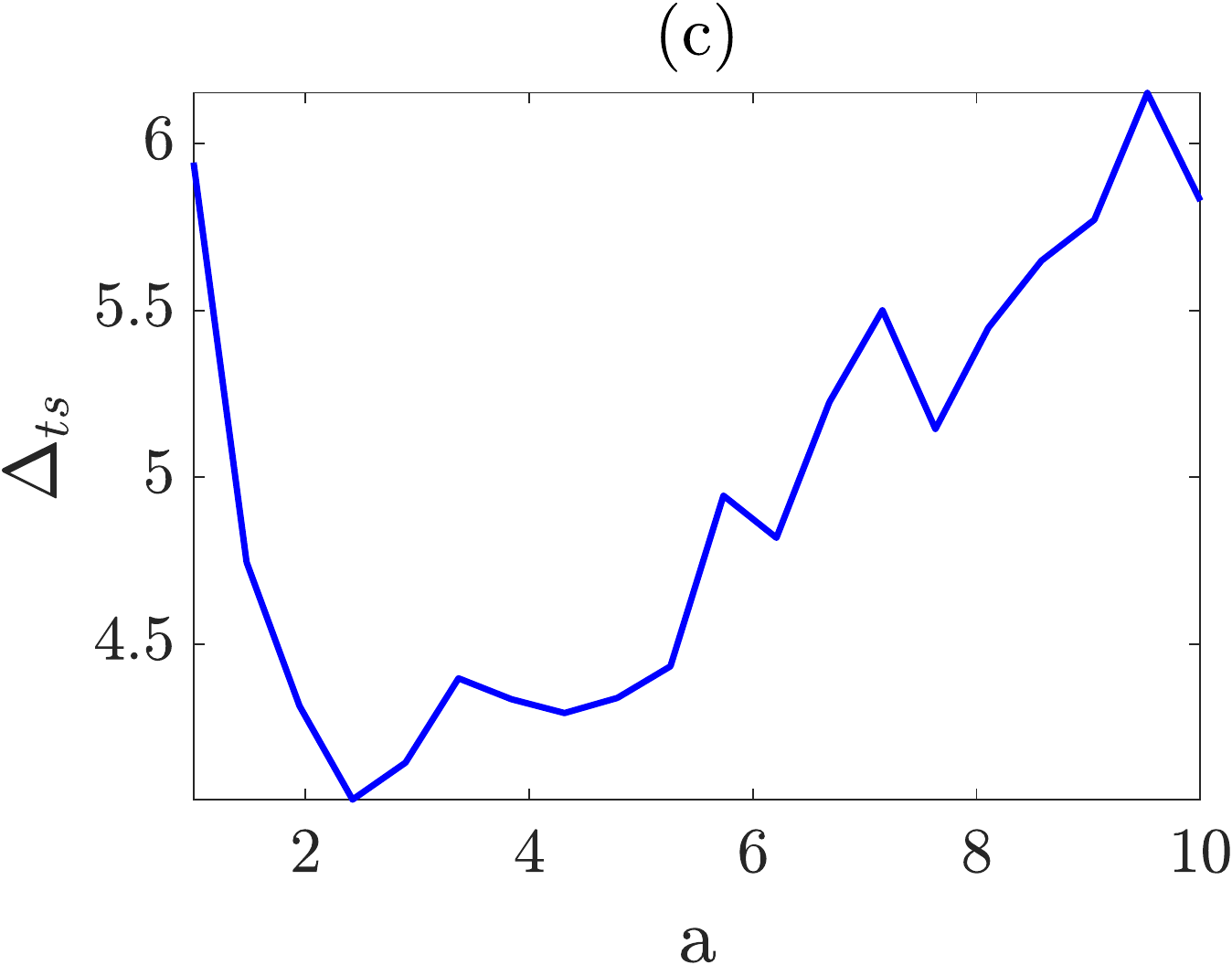}
\caption{Testing errors for the (a) Lorenz, (b) Roessler and (c) HR systems, plotted as a function of the LPF cutoff frequency $a$.}
 \label{fig:Error_vs_a}
\end{figure}

{In practice, one may only have access to the input signal corrupted with noise, and the underlying chaotic signal may not be known a \textit{priori}. This prevents computation of the optimal cutoff frequency, hence it may become necessary to drive the input signal through the LPF while varying the cutoff frequency `$a$'. To assess the effects of the LPF, we consider the case where the noise strength of the training and testing signals are fixed, i.e., $\epsilon_1=5$, $\epsilon_2=20$}. Figure \ref{fig:Error_vs_a} shows the testing error plotted against `$a$'. We see that the minimum testing error is obtained for the above-mentioned values of $\widehat{a}$ associated with each chaotic system. That is, for the Lorenz system the minimum $\Delta_{ts}$ is obtained when the cutoff frequency is around $a=13$. Similarly, for the Roessler system and the Hindmarsh-Rose system, the minimum $\Delta_{ts}$ is obtained when $a=2$ and $a=3$, respectively. 

\begin{figure} [htb!]
 \centering
 \includegraphics[scale=0.4]{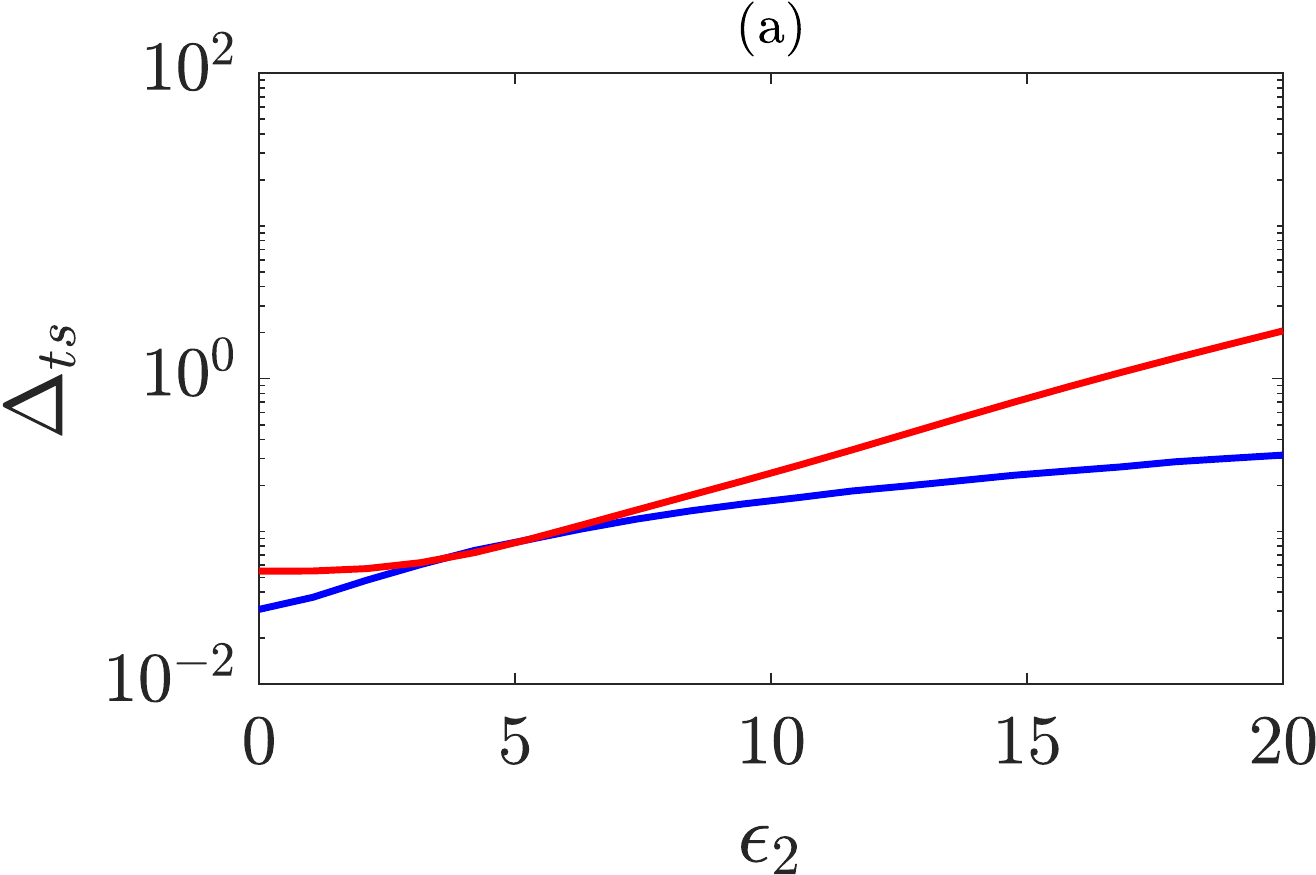}
 \includegraphics[scale=0.4]{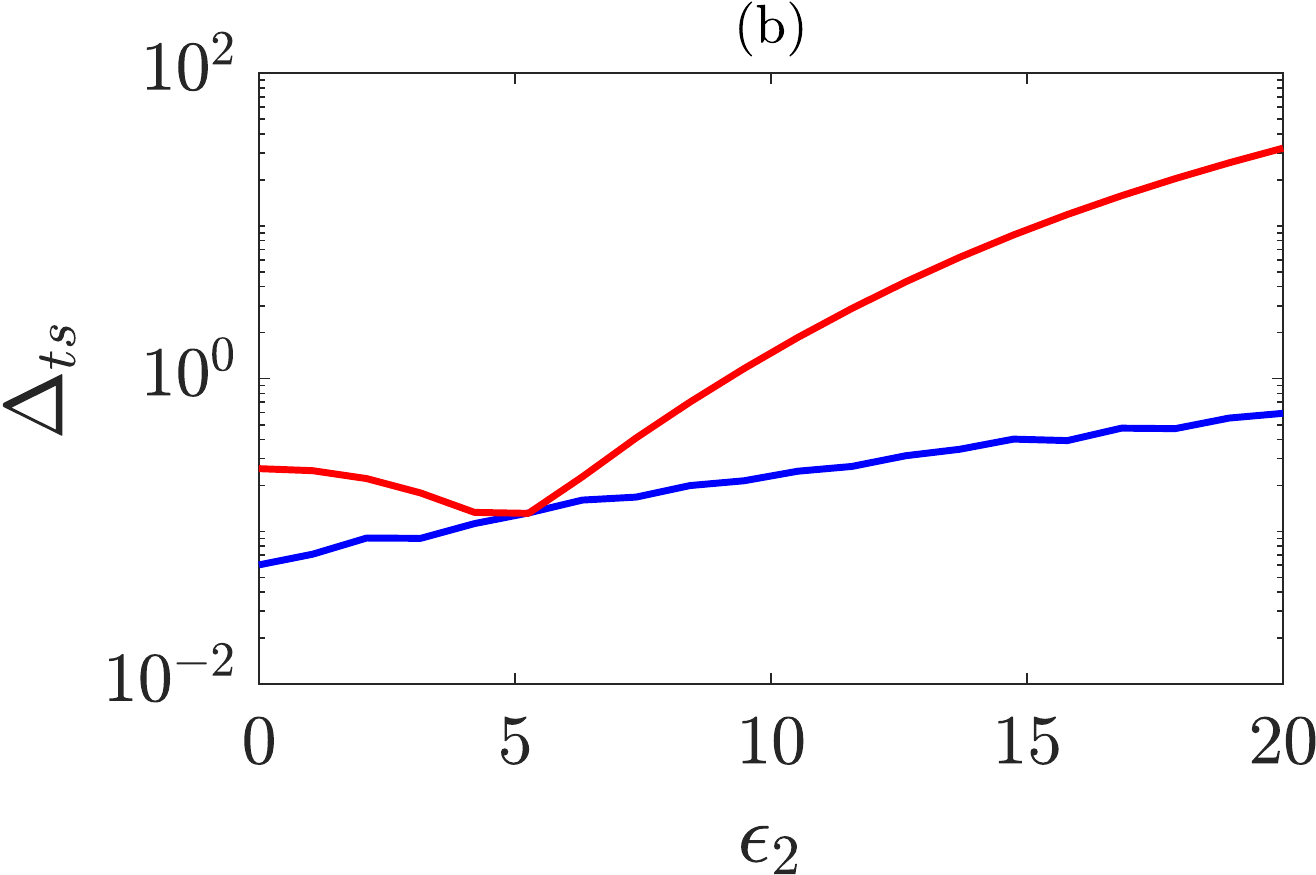}
 \includegraphics[scale=0.4]{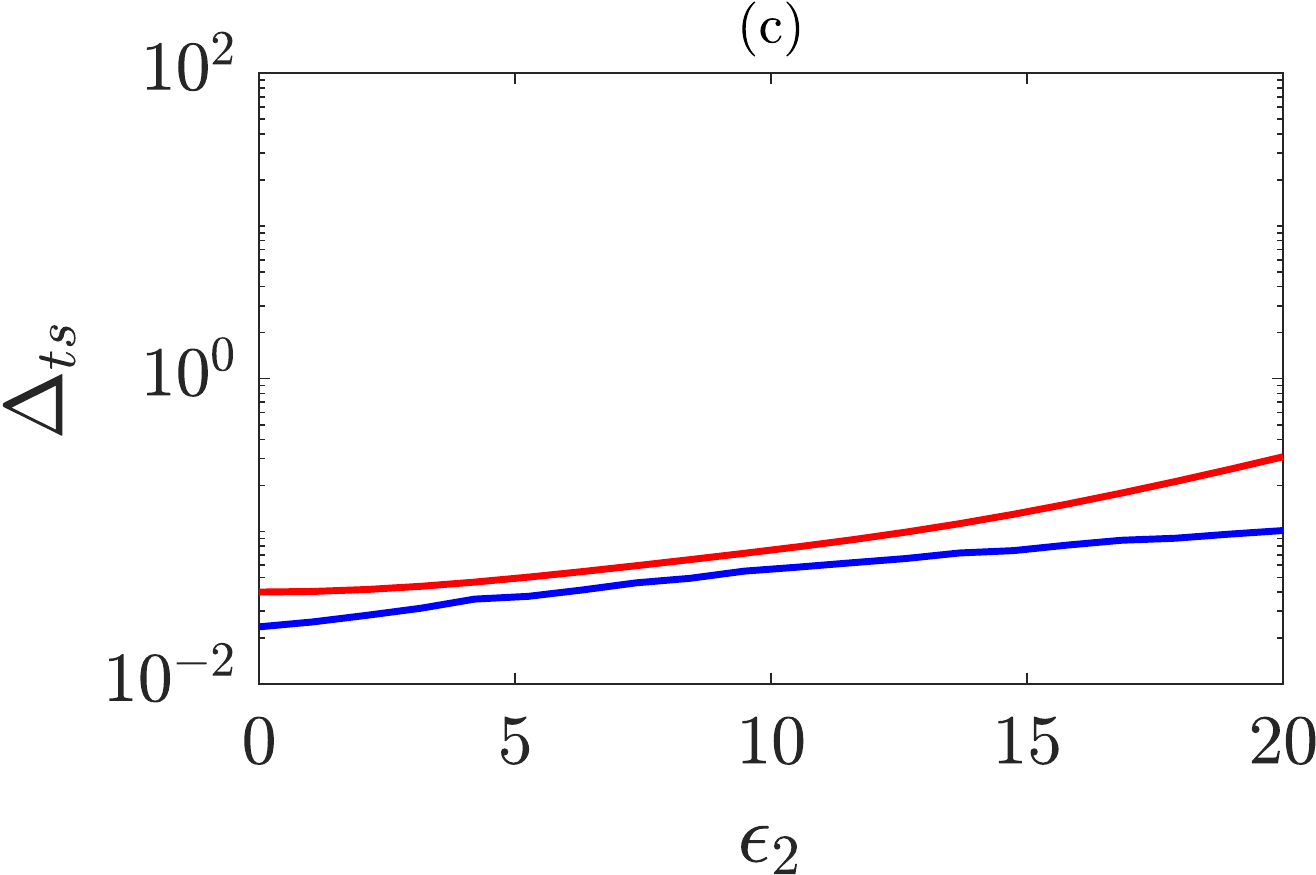}
 \caption{Testing errors for the (a) Lorenz, (b) Rossler, and (c) Hindmarsh-Rose systems.The blue color plot is the error when the input signal is driven through the LPF and red color plot is error when the input is not driven through the LPF.}
 \label{fig:Error_vs_epsilon}
\end{figure}

Next, we investigate the effect of varying the noise strength. Namely, we fix the noise strength of the training phase $\epsilon_1=5$ and vary $\epsilon_2$. For each system we set the cutoff frequency equal to $\widehat{a}$. Figure \ref{fig:Error_vs_epsilon} shows that, for all cases, the testing error is lower when the input signal is driven through the LPF, compared to the case when the filtering is not applied. This is true also in the case that no noise is added to the input signal in the testing phase, i.e., $\epsilon_2=0$.  
The training error is independent of the LPF as that is independent of $\epsilon_2$, and we find that the training error for the filtered case is lower than for the unfiltered case.


\begin{figure}[H]
    \centering
    \includegraphics[scale=1.25]{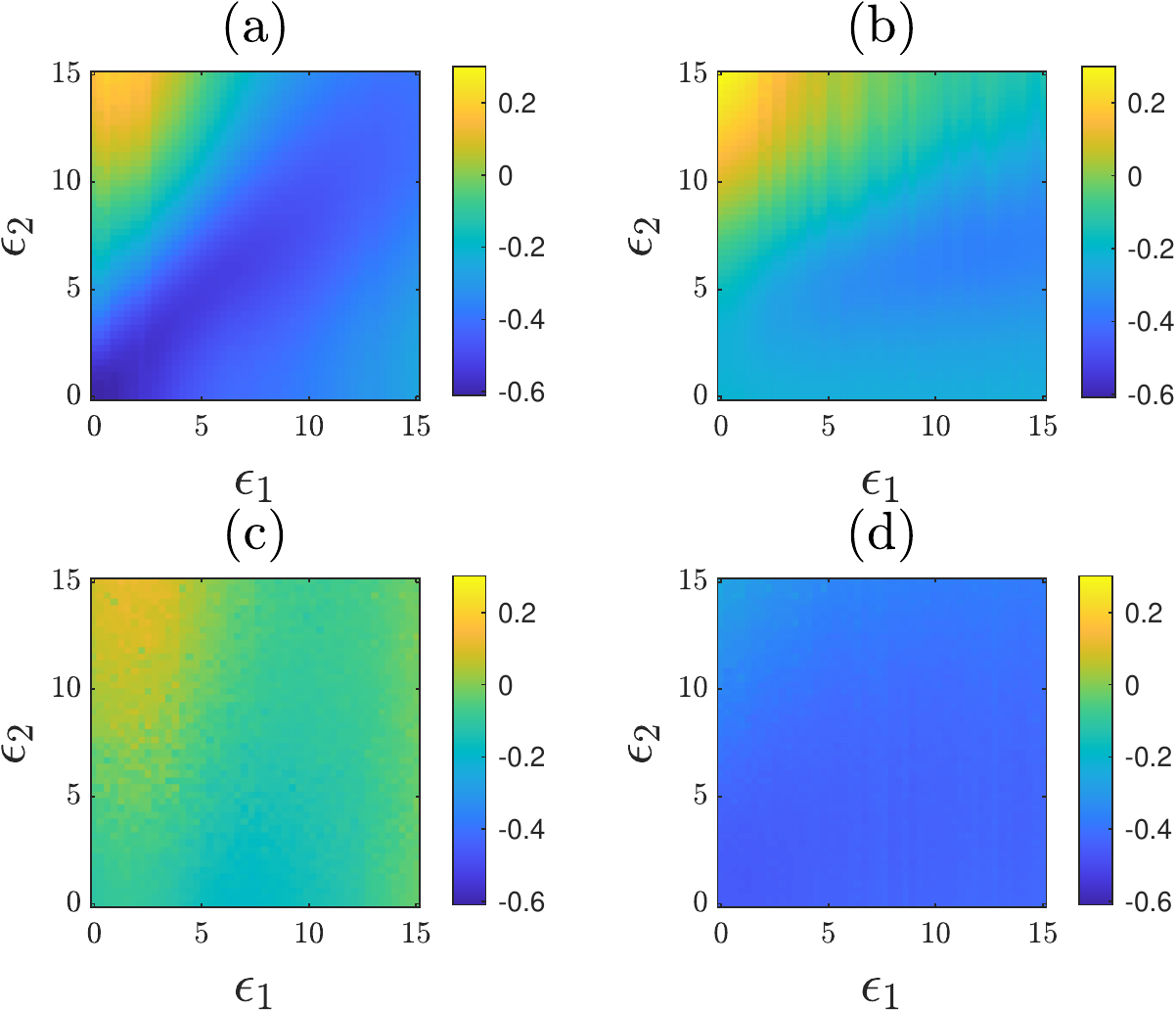}
    
    \caption{Contour plots of the  $\log_{10}$  of the testing error for the case of the Lorenz task and different cases: (a)  the LPF is not used (b) only the input signal in the training phase $\tilde{s}^{tr}(t)$ is driven through the LPF, (c) only the input signal in the testing phase  $\tilde{s}^{ts}(t)$ is driven through the LPF and (d) both signals $\tilde{s}^{tr}(t)$ and $\tilde{s}^{ts}(t)$ are driven through the LPF. }
     \label{fig:contours_filtered}
\end{figure}

We assessed the performance of reservoir computing by filtering the training and testing input signals individually. Figure  \ref{fig:contours_filtered} shows contour level plots of the testing error for the Lorenz system task in various cases. In case (a), the error is computed without filtering the training or testing signals. { Panel (a) of Fig.\ \ref{fig:contours_filtered} is the same as Panel (a) of Fig.\  \ref{fig:contours}, except we changed the color legend to be consistent with the other plots in Fig.\ \ref{fig:contours_filtered}} {A LPF with cutoff frequency $\widehat{a} = 12$ is used for the rest of the cases. In case (b), the reservoir is trained with the filtered version of $\tilde{s}_{\epsilon_1}^{tr}(t)$, i.e., $\hat{s}_{\epsilon_1}^{tr}(t)$ (but the input signal in the testing phase is still $\tilde{s}_{\epsilon_2}^{ts}(t)$). This indicates that noise is removed only from the input signal in the training phase (but not from the input signal in the testing phase). { As a result, the performance of the reservoir is more robust to changes in $\epsilon_1$ (especially for low values of $\epsilon_2$) than to changes in $\epsilon_2$.} For lower values of $\epsilon_2$, since the input signal in the training phase is filtered the testing error is low. 
In case (c), the reservoir is trained with the noise-corrupted version of the input signal in the training phase, i.e., $\tilde{s}_{\epsilon_1}^{tr}(t)$, whereas, noise is removed only from the input signal in the testing phase, i.e. $\hat{s}_{\epsilon_2}^{ts}(t)$. {We observed that when $\epsilon_1$ is small, a small amount of overfitting is obtained. However, as $\epsilon_1$ increases, the training signal becomes too noisy irrespective of the regression parameter and overfitting. Lastly,} we consider the case where both the $\tilde{s}_{\epsilon_1}^{tr}(t)$ and $\tilde{s}_{\epsilon_2}^{ts}(t)$ are filtered. Since the frequencies above $\widehat{a}$ are rejected, the impact of noise on the reservoir computer is effectively reduced. Consequently, we obtain a very low $\Delta_{ts}$ over the entire $\epsilon_1,\epsilon_2$ plane, which is shown in case (d). We conclude that when no filtering operation is applied the best performance is obtained for $\epsilon_1=\epsilon_2$. However, for $\epsilon_1 \neq \epsilon_2$ the reservoir performance can be improved by filtering both the input signal in the training phase and in the testing phase with the same cutoff frequency  $\widehat{a}$. }

We also considered the effects of picking different cutoff frequencies of LPF applied to the input signals in the training phase and in the testing phase. This is discussed in Sec.\ 2 of the Supplementary Material.

\section{Conclusion}

This paper presents a  comprehensive investigation of the use of a low-pass filter to mitigate the effects of noise on the performance of a reservoir observer. 
The effects of noise in both the training phase and in the testing phase were considered, and for both the cases that noise affects the input signals and the training and testing signals. Overall, low-pass filters are found to provide a good remedy against noise.

We first consider the case that filtering is not applied and find that the best performance is achieved when the noise strength affecting the input signal is about the same in the training and in the testing phases. The performance in the case that the amount of noise is the same is higher than when the reservoir is trained with noise and tested in a noise-free environment. This motivates us to study possible remedies to implement when these amounts are not the same. 
We thus introduce low-pass-filtering applied to the input signals both in the training phase and in the testing phase. We  investigate the performance of the RC as the cutoff frequency of the low pass filter is varied and find the optimal value of the cutoff frequency.
We see a substantial improvement in the testing error, provided that the same type of filtering is applied in the training phase and in the testing phase.


One conclusion that we obtain is that it may be good to filter input and output signals, even when an estimate on the amount of noise that affects these signals is not available. In fact, low-pass filtering is typically not found to be detrimental, even when the signals are noise free. However, a more significant improvement in performance is observed when a low pass filter is applied to signals affected by increasing amount of noise.

\section*{Supplementary Material}
{
The supplementary material includes information about the different tasks that we assign to a reservoir computer in this paper {and a study of the performance of a reservoir computer that uses different cutoff frequencies in the training phase and in the testing phase.}}

\section*{Acknowledgement} This work was partly funded by NIH Grant No. 1R21EB028489-01A1 and by the Naval Research Lab’s Basic Research Program.
\section*{Data Availability}
The data that support the findings of this study are available
within the article.

\bibliographystyle{aipnum4-1}
\bibliography{noisy_rc}

\end{document}


\title{Supplementary Information for\\ Reservoir Computing with Noise}
\author{Chad Nathe}
	\affiliation{Mechanical Engineering Department, University of New Mexico, Albuquerque, NM, 87131}
\author{Chandra Pappu}
	\affiliation{Electrical, Computer and Biomedical Engineering Department, Union College, Schenectady, NY, 12309}
\author{Nicholas A. Mecholsky}
	\affiliation{Department of Physics and Vitreous State Laboratory, 
The Catholic University of America,
Washington, DC 20064}
\author{Joe Hart}\affiliation{US Naval Research Laboratory, Washington, DC 20375}
		\author{Thomas Carroll}
	\affiliation{US Naval Research Laboratory, Washington, DC 20375}
\author{Francesco Sorrentino}
	\affiliation{Mechanical Engineering Department, University of New Mexico, Albuquerque, NM, 87131}

	\maketitle

\section{Description of different tasks}

 The underlying process we want to model may evolve in time based on a set of deterministic (chaotic) equations, such as the equations of the Lorenz chaotic system, in the variables $x(t), y(t), z(t)$ (See Eq.\ \eqref{Lorenz system}.)
One task that can be given to the RC is to reconstruct the time evolution of the training signal, e.g. $y(t)$ from knowledge of the input signal, e.g., $x(t)$. In what follows we describe different tasks that will be given the reservoir, which in the main text we simply refer to as the `Lorenz task', the `Roessler task' and the 'Hindmarsh Rose (or HR) task'. For each one of these tasks, the input signal $s(t)$ is chosen to coincide with one of the state variables of the chaotic system and the output signal with another state variable from the same system.

\noindent
The Lorenz chaotic system is modeled by the following set of equations,
\begin{equation} \label{Lorenz system}
    \begin{aligned}
	    \dot x(t) & = 10(y(t)10-x(t)) \\
		\dot y(t) & = x(t)(28-z(t))-y(t) \\
		\dot z(t) & = x(t)y(t) - 8/3 z(t)
    \end{aligned}
\end{equation}
For this task, the $x(t)$ state variable is used as the input signal $s(t)$. The $z(t)$ state variable is used as the training signal $g(t)$. \\
The Roessler chaotic system is modeled by the following set of equations,
\begin{equation} \label{Rossler system}
    \begin{aligned}
	    \dot x(t) & = -y(t) - z(t) \\
		\dot y(t) & = x(t)+0.15 y(t)\\
		\dot z(t) & = 0.2+ x(t)z(t) -4 z(t).
    \end{aligned}
\end{equation}
For this task, the $x(t)$ state variable is used as the input signal $s(t)$. The $z(t)$ state variable is used as the training signal $g(t)$.

\noindent The Hindmarsh-Rose (HR) chaotic system is modeled by the following set of equations,
\begin{equation} \label{Hindmarsh-Rose system}
    \begin{aligned}
	    \dot x (t) & = y (t) + \phi [x (t)] - z (t) + 3.2\\
	    \dot y (t) & = \psi [x (t)] -y (t)\\
	    \dot z (t) & = p(4(x (t) + 8/5) - z (t))\\ 
	\end{aligned}
\end{equation}  
where,
\begin{equation*} 
	\begin{aligned}  
	    \phi [x(t)] &= -x^3 + 3x^2\\
	    \psi [x(t)] & = 1 - 5x^2\\
    \end{aligned}
\end{equation*}
and $p= 0.006$. 

For this task, the $x(t)$ state variable is used as the input signal $s(t)$. The $z(t)$ state variable is used as the training  signal $g(t)$.

For all tasks the sampling time $t_s=0.01$.

\section{Effects of low-pass filtering with different cutoff frequencies}

\begin{figure} [H]
 \centering
 \includegraphics[scale=0.65]{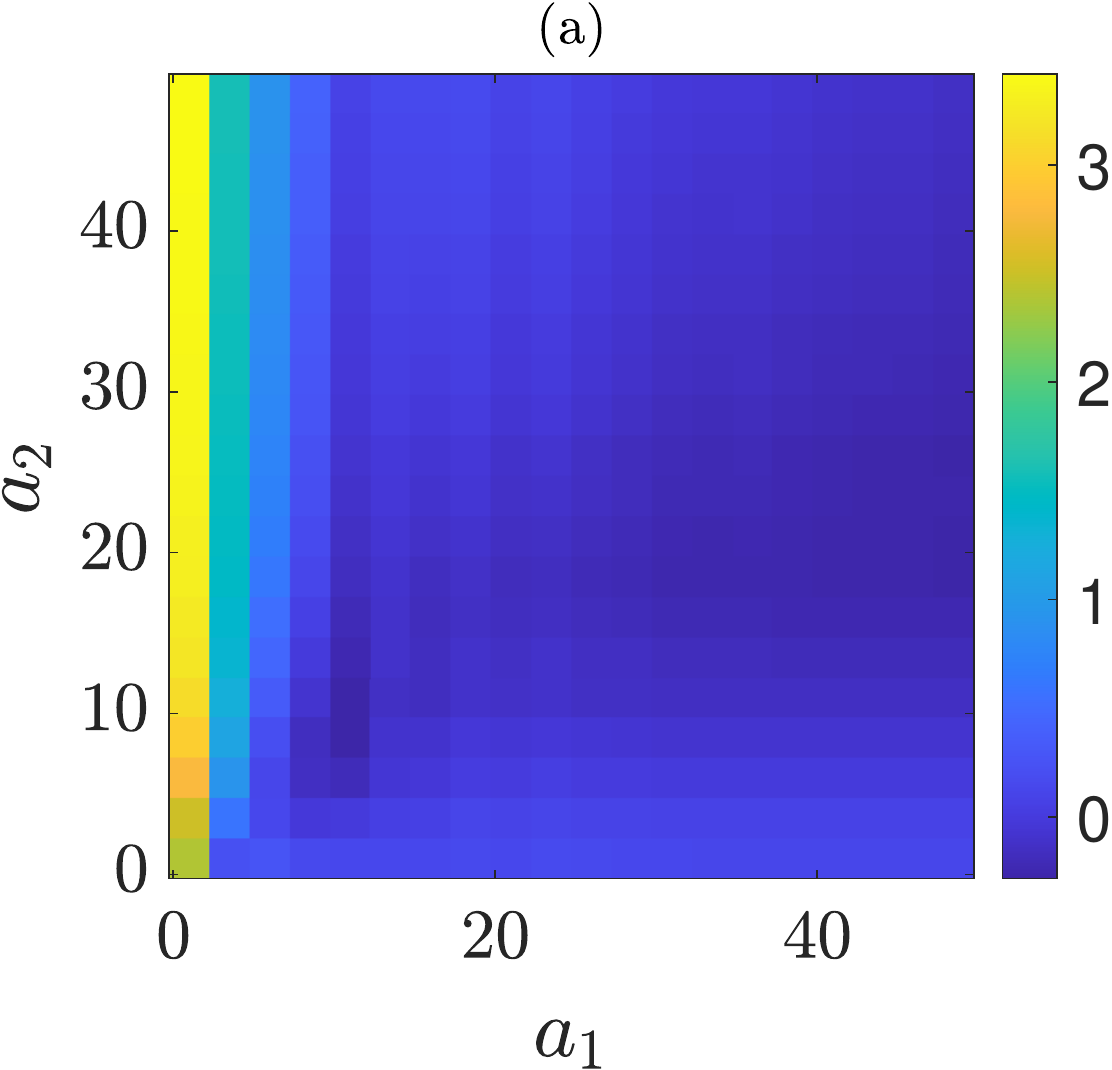}
\caption{Testing errors for the Lorenz system plotted against $a_1$ and $a_2$.}
 \label{fig:Error_a1_vs_a2}
\end{figure}

We also studied the behavior of the reservoir computer by considering two low pass filters with cutoff frequencies $a_1$ and $a_2$. For this case, we fixed $\epsilon_1=5$ and $\epsilon_1=20$. The first LPF is used to filter the noise corrupted versions of the input training signal $\tilde{s}^{tr}(t)$ with cutoff frequency $a_1$ and the second LPF is used to filter the noise corrupted versions of the input training signal $\tilde{s}^{tr}(t)$ with cutoff frequency $a_2$.
Since the optimal cutoff frequency of the LPF used for the Lorenz system, $\widehat{a}$ is around 12,  as discussed in the main manuscript, the minimum testing error $\Delta_{ts}$ should be obtained around $a_1=a_2=12$. Figure \ref{fig:Error_a1_vs_a2} shows the contour plot of the $\Delta_{ts}$ plotted as a function of $a_1$ and $a_2$. The highest testing error $\Delta_{ts}$ is obtained when $a_1=1$, that is, the input training signal $\tilde{s}^{tr}(t)$ is completely filtered thereby changing the dynamics of the chaotic signal. However, as expected the minimum $\Delta_{ts}$ is obtained around optimal cutoff frequency $\widehat{a}_1=\widehat{a}_2=12$ and the reservoir computer has improved performance when both the $\tilde{s}^{tr}(t)$ and $\tilde{s}^{ts}(t)$ are filtered. 

\section{Effects of the Network Topology}

In this section, we experiment with different network topologies and observe the effect on the training and testing errors. We start with a Erdos-Renyi networks with connectivity probability, $p$, which we can take as either undirected or directed. We perform the operation, $A \leftarrow A/ \rho(A)$, where $\rho(A)$ is the spectral radius so that the largest eigenvalue of the normalized matrix has modulus equal to $1$. In Fig. 2, we observe that the training and testing errors are completely independent of whether or not the network is undirected. We also see that the effect of changing $p$ is not drastic unless $p$ is large. We experiment with systems with noise and without noise. {For the system where the input signal is corrupted with noise, we considered $\epsilon_1=\epsilon_2=5$. The difference in training and testing errors between undirected and directed networks is negligible for both the noise-free and noise-corrupted cases. However, as the input is noise corrupted, the errors tend to be high compared to the noise-free input signal. Irrespective of the case, as $p$ is increased, the training and testing errors gradually increase. Therefore, the results shown in this work hold for any case of $p$ unless it is large. As it can be observed that for the large values of $p$, i.e., $p>0.9$, the errors are substantial for both the undirected and directed networks, and hence, these values of $p$ should be avoided.}

\begin{figure}[H]
    \centering
    \includegraphics[width=.45\textwidth]{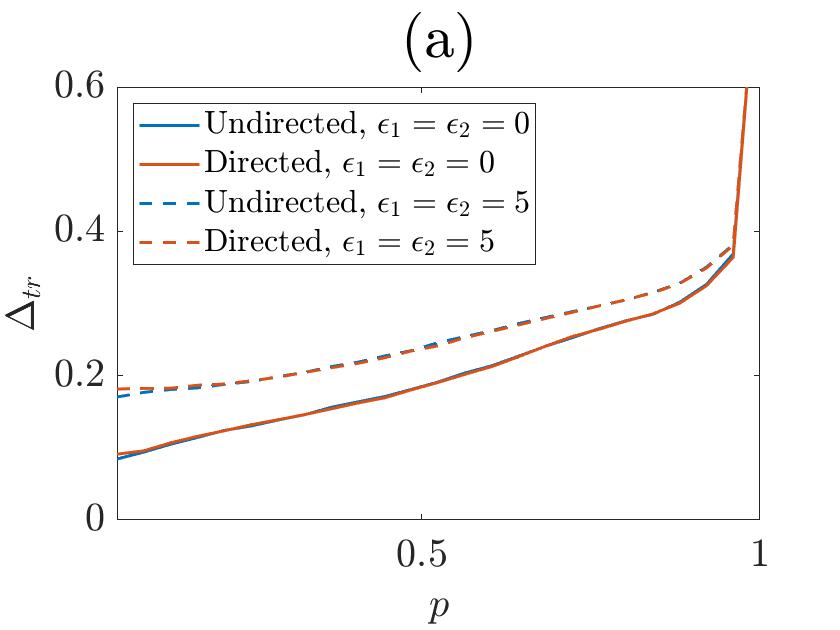}
    \includegraphics[width=.45\textwidth]{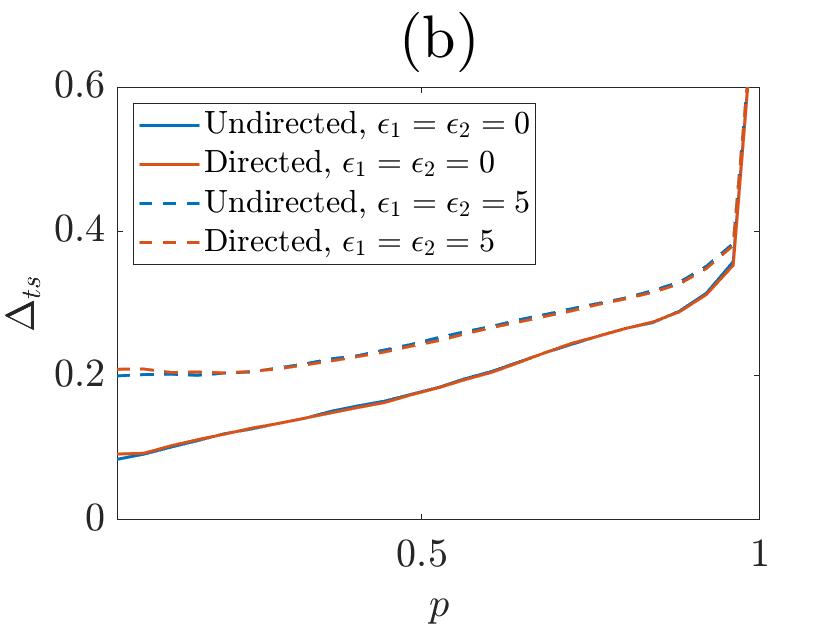}
    \caption{We plot the training and testing errors versus connected probability, $p$, for both undirected and directed networks. We do so in the case of no noise ($\epsilon_1=\epsilon_2=0$) as well as in the case of noise ($\epsilon_1=\epsilon_2=5$). The $x$-axis starts at $p=0.05$ (a) Training error (b) Testing error.}
\end{figure}

\section{Optimization of $\alpha$}

In Fig. 3, we report the testing error as a function of leakage rate controlling parameter, $\alpha$, for the three chaotic tasks shown in the paper. We show how $\alpha$ affects the testing error for both the cases of no noise and with noise. {For the case of noise, we consider $\epsilon_1=\epsilon_2=5$. For all the systems, the error plotted against $\alpha$ is consistent for both the cases. As expected, the testing error is high in the noise-corrupted case. The minimum testing errors for the Lorenz, the Rossler and the HR systems are obtained at $\alpha = 0.1$, $\alpha=0.003$ and $\alpha=0.01$, respectively. And hence, the corresponding $\alpha$'s are considered as optimum values which are used throughout the paper. The reservoir computer that uses $\alpha$ below and above the optimized value yields significant error values.} 

\begin{figure}[H]
    \centering
    \begin{tabular}{l l l}
    \text{(A)} & \text{(B)} & \text{(C)}\\
    \includegraphics[width=.33\textwidth]{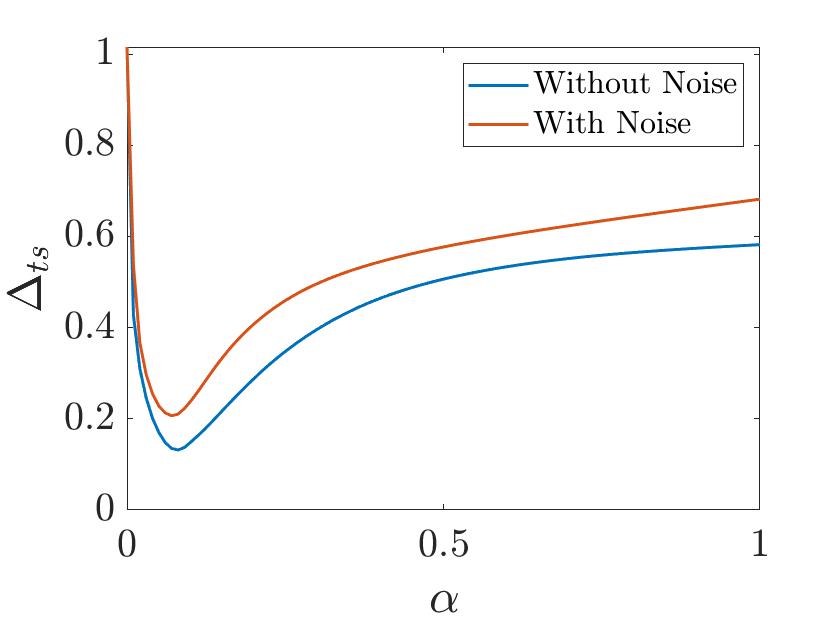} &
    \includegraphics[width=.33\textwidth]{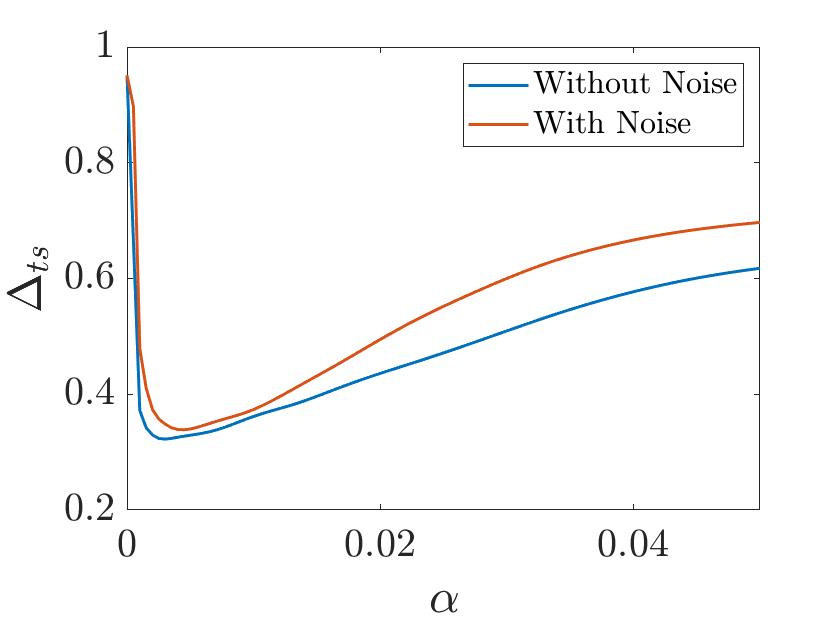} &
    \includegraphics[width=.33\textwidth]{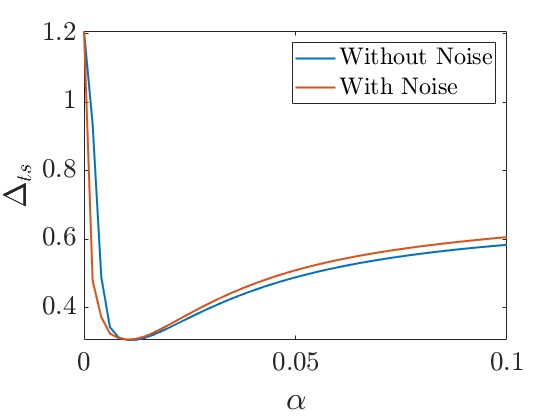}
    \end{tabular}
    \caption{We plot the testing error versus the linearity controlling parameter, $\alpha$. We average over several random choices of $A$ and $\mathbf{w}$. We plot both the instances of no noise, as well as noise given by $\epsilon_1=\epsilon_2=5$, $\epsilon_3=\epsilon_4=0$. (A) Lorenz (B) Rossler and (C) HR systems.}
\end{figure}